\documentclass{article}
\usepackage{spconf,amsmath,graphicx}

\usepackage{enumitem}
\setlist{nosep, leftmargin=14pt}

\usepackage{mwe} 

\title{SimSAM: Zero-shot Medical Image Segmentation \\ via Simulated Interaction}
\name{Benjamin Towle$^{\star}$ \qquad Xin Chen$^{\star}$ \qquad Ke Zhou$^{\star \dagger}$}
\address{$^{\star}$ School of Computer Science, University of Nottingham \\  $^{\dagger}$ Nokia Bell Labs \\ \texttt{\{firstname.lastname\}@nottingham.ac.uk}}
\usepackage{graphicx}
\usepackage{amsmath}
\usepackage{amssymb}
\usepackage{booktabs}
\usepackage{xcolor}

\usepackage{algorithm}
\usepackage{algpseudocode}
\usepackage{url}

\usepackage{breqn}
\usepackage{multicol}
\usepackage{multirow}
\DeclareMathOperator*{\simm}{sim}

\DeclareMathOperator*{\iou}{IoU}

\DeclareMathOperator*{\abs}{abs}

\DeclareMathOperator*{\argmax}{argmax}

\DeclareMathOperator*{\topk}{TopK}

\begin{document}
%
\maketitle
\begin{abstract}
The recently released Segment Anything Model (SAM) has shown powerful zero-shot segmentation capabilities through a semi-automatic annotation setup in which the user can provide a prompt in the form of clicks or bounding boxes. There is growing interest around applying this to medical imaging, where the cost of obtaining expert annotations is high, privacy restrictions may limit sharing of patient data, and model generalisation is often poor. However, there are large amounts of inherent uncertainty in medical images, due to unclear object boundaries, low-contrast media, and differences in expert labelling style. Currently, SAM is known to struggle in a zero-shot setting to adequately annotate the contours of the structure of interest in medical images, where the uncertainty is often greatest, thus requiring significant manual correction. To mitigate this, we introduce \textbf{Sim}ulated Interaction for \textbf{S}egment \textbf{A}nything \textbf{M}odel (\textsc{\textbf{SimSAM}}), an approach that leverages simulated user interaction to generate an arbitrary number of candidate masks, and uses a novel aggregation approach to output the most compatible mask. Crucially, our method can be used during inference directly on top of SAM, without any additional training requirement. Quantitatively, we evaluate our method across three publicly available medical imaging datasets, and find that our approach leads to up to a 15.5\% improvement in contour segmentation accuracy compared to zero-shot SAM. Our code is available at \url{https://github.com/BenjaminTowle/SimSAM}.

\end{abstract}
\section{Introduction}
\label{sec:intro}
Large pre-trained foundation models that exhibit powerful zero-shot generalisation through careful prompt design, without requiring parametric re-training, are increasingly becoming the \textit{de facto} approach across numerous fields in machine learning \cite{Bommasani2021OnTO, Brown2020LanguageMA, Rombach2021HighResolutionIS}. In image segmentation, the recently released Segment Anything Model (SAM) \cite{sam} has demonstrated state-of-the-art semi-automatic zero-shot capabilities, through re-framing the prompt as interaction information from the user such as clicks, bounding boxes or masks, which guide the annotation of the region of interest \cite{hai_imis}. 

There is growing interest in applying SAM to Medical Image Segmentation \cite{medsam, medsam-mcl, medsam-zs, Mohapatra2023SAMVB}. Previous approaches to medical imaging focused on supervised training of bespoke models for each task, requiring numerous manually labelled examples \cite{medsam-zs}. Yet, this presents several limitations: the requirement for a trained clinician renders annotation costs extremely high \cite{Rahimi2021AddressingTE}; further, privacy restrictions may prevent sharing of patient data \cite{Adnan2021FederatedLA}, limiting the availability of the kind of large-scale datasets seen in other domains such as NLP; finally, these models often show poor generalisability out of the lab, e.g. due to variability in modality or device used to obtain the images \cite{Ali2022AssessingGO}. Resultantly, SAM could greatly speed up existing clinical pipelines, by enabling rapid semi-automatic segmentation of medical images \cite{medsam-zs}. 

\begin{figure}
    \centering
    \includegraphics[scale=0.11]{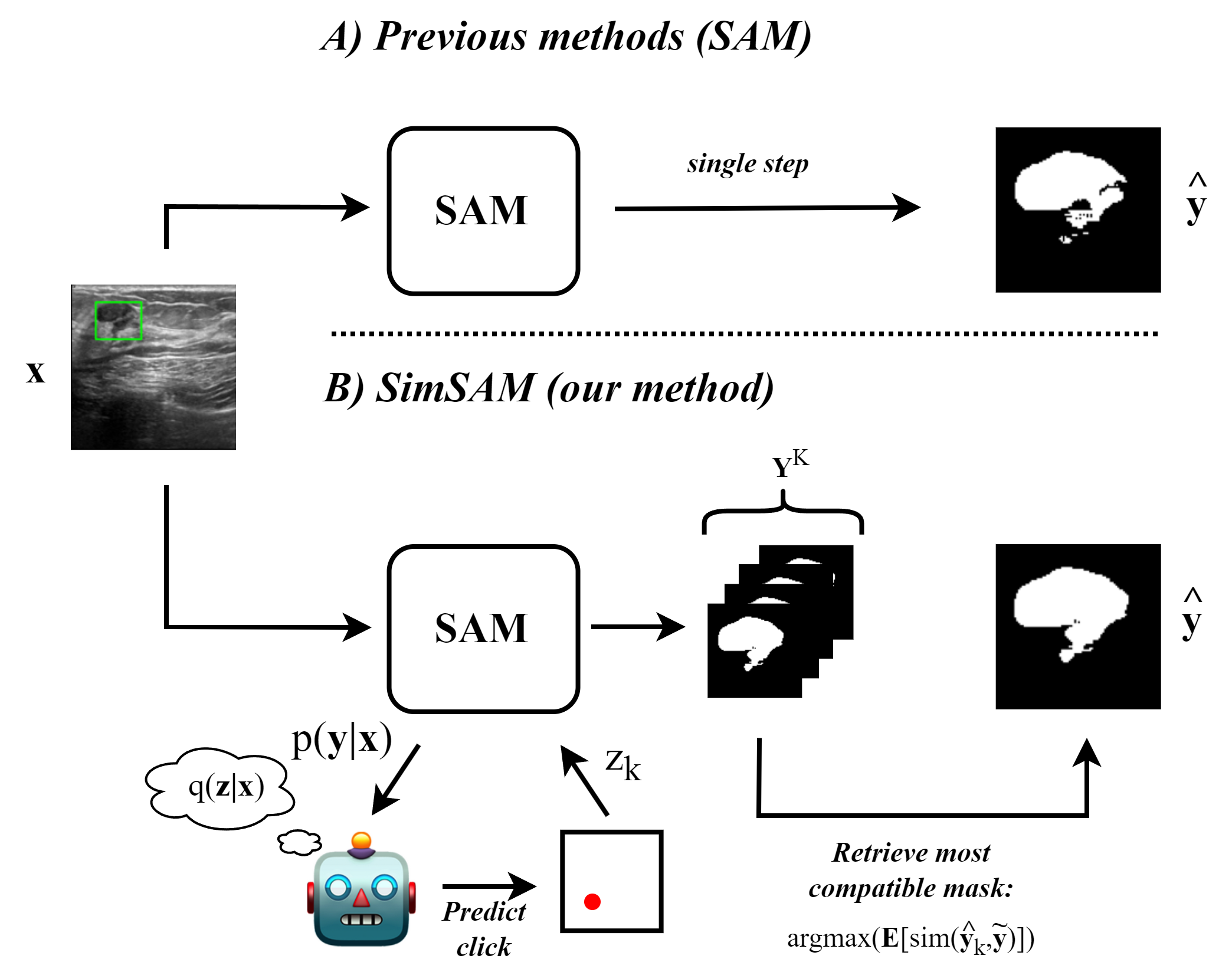}
    \caption{(A) Previous methods predict annotation mask in a single step; (B) \textsc{SimSAM}, our method, simulates possible click locations to provide an additional prompt to the model and retrieves the most compatible mask from a pool of generated masks. Example is from Breast Ultrasound Scan test set \cite{AlDhabyani2019DatasetOB}.}
    \label{fig:simsam}
\end{figure}

However, unlike natural images, medical images often have significant uncertainty around their contours, e.g. due to unclear boundaries between healthy and unhealthy tissue, or noise from low-contrast media. This leads to substantial expert disagreement about an image's correct annotation \cite{Kohl2018APU}. Appropriately determining these contours has major downstream consequences, such as deciding how invasive a surgery will be in the case of a tumour \cite{Monteiro2020StochasticSN}.  When attempting this as a zero-shot task, SAM is known to perform particularly poorly at appropriately segmenting the edges of the image \cite{medsam-zs}, resulting in significant additional manual corrections being required by the clinician to obtain satisfactory accuracy.

We observe that unlike segmentation models which produce a dense pixel-wise output in a single forward pass, a human clinician creates annotations in more of a sequential process, enabling more careful consideration between alternate hypotheses for the contours. Concurrently, emergent capabilities have been discovered in many foundation models through chain-of-thought prompting \cite{Wei2022ChainOT} -- i.e. feeding the model's outputs back into itself as a prompt, to enhance its final generation. This enables the model to draw upon `dark knowledge' \cite{Hinton2015DistillingTK} -- information learned during training, but not immediately visible in its outputs -- to guide its predictions. Intuitively, this allows a model to break a problem down into incremental steps, rather than requiring it to make an immediate prediction. So far however, there has been a lack of work exploring how this principle can be applied to interactive segmentation models, whose `prompts' are clicks and bounding boxes, rather than text.  

In this paper, we introduce \textbf{Sim}ulated Interaction for \textbf{S}egment \textbf{A}nything \textbf{M}odel (\textbf{\textsc{SimSAM}}), a zero-shot extension to SAM that significantly enhances SAM's out-of-the-box performance, without requiring any additional training. Specifically, \textsc{SimSAM} uses a carefully-designed click simulation mechanism that extracts knowledge about locations of user clicks to form additional prompts to enhance its predictions. We further propose a method for aggregating these predictions, which maximises the compatibility of the outputted images. Figure \ref{fig:simsam} demonstrates how in contrast to vanilla SAM which predicts the mask in a single step, \textsc{SimSAM} is able to iteratively improve its prediction through producing multiple masks from self-generated prompts, and aggregating over these masks. We evaluate our method across three publicly available medical imaging datasets. Quantitatively, we find that our approach consistently outperforms SAM across all three datasets, with up to 15.5\% improvement in contour segmentation accuracy. Qualitatively, we demonstrate the superior annotations of our approach, and show how our approach is able to generate more robust masks that mitigate many of the pitfalls of SAM.

\section{Method}
Figure \ref{fig:simsam}B overviews our method. Given an input image of $N$ pixels $\textbf{x} = \{x_n \}_{n=1}^N$, our goal is to predict a binary mask $\mathbf{\hat{y}}$, that matches an expert annotation $\mathbf{y}$, using the Segment Anything Model (SAM). We first show how $\mathbf{\hat{y}}$ can be predicted by marginalising over a known distribution of user clicks, then show how this process can be approximated by SAM.

\subsection{Segment Anything Model}
SAM is a foundation segmentation model trained on over 1B masks from 11M images \cite{sam}, with the ViT backbone transformer encoder \cite{vit}. After encoding an image, the model enables an output mask to be iteratively refined through conditioning on various `prompts', e.g. user clicks and bounding boxes, which are attended to by the model's decoder.

\subsection{Marginalising over Prompts}
We assume that each mask $\textbf{y}$ is conditioned not only on the input image $\textbf{x}$ but also on a prompt $\textbf{z}$ whose possible values represent the available points a user might click $\{ z_n \}_{n=1}^N$. The ground-truth probability distribution over user clicks is then given by $p(\textbf{z} \vert \textbf{x})$. Given input image $\textbf{x}$, SAM first produces a dense pixel-wise probability distribution $p(\mathbf{y} \vert \textbf{x})$. Then, we sample a click from $p(\textbf{z} \vert \textbf{x})$ and make a new prediction $\mathbf{\hat{y}}$ conditioned on this click. Thus, we estimate the probability for the output mask, by marginalising over user clicks as follows:
\begin{equation}
\label{eq:eiou}
    p(\textbf{y} \vert \textbf{x}) = \sum_n^N p(\textbf{y} \vert \textbf{x}, z_n)p(\textbf{z} = z_n \vert \textbf{x})
\end{equation}
\subsection{Simulating User Clicks}
We do not in practice have access to a ground-truth distribution $p(\mathbf{z} \vert \textbf{x})$, and requiring a user to provide this would defeat the purpose of improving SAM's performance without requiring any additional manual human annotation. Instead, we would like to obtain some distribution $q(\mathbf{z} \vert \textbf{x})$ that approximates this. Let $\textbf{e} = \mathbf{fp} \cup \mathbf{fn}$ be the error mask of incorrectly annotated pixels, comprising the union of false positives $\mathbf{fp}$ and false negatives $\mathbf{fn}$. We follow the assumption from previous work that the user will click one of these pixels \cite{Liu2022PseudoClickII}. Then, although the model is not explicitly trained to predict an error mask, we observe that the model can implicitly provide a zero-shot approximation of the probabilities for these, simply by transforming the original probability mask $p(\mathbf{y} \vert \textbf{x})$, namely:
\begin{equation}
    p(e_n = 1) = 0.5 - \abs(p(y_n \vert \textbf{x}) - 0.5)
\end{equation}
We emphasise that this click simulation is inferred entirely from the zero-shot probabilities of SAM, \textit{without} requiring any additional gradient updates to the model.

\subsection{Top K Approximation}
\label{sec:topk}
Even approximating $p(\mathbf{z} \vert \textbf{x})$ with $q(\mathbf{z} \vert \textbf{x})$, it is intractable to calculate Equation (\ref{eq:eiou}) exactly, as the number of possible clicks is equal to the number of pixels in the image $N$. We therefore approximate this through taking the top K clicks:
\begin{equation}
    p(\textbf{y} \vert \textbf{x}) \approx \frac{1}{K} \sum_k^K p(\textbf{y} \vert \textbf{x}, z_k), z_k \in \topk(p(\mathbf{z} \vert \mathbf{x}))
\end{equation}
\subsection{Image-level Aggregation}
\label{sec:image-level}
While the above method enables marginalisation over pixel-level outputs, it does not explicitly consider the interdependencies between pixels. Simply independently averaging pixel-values across each of the masks may fail to produce a mask that is overall coherent. We therefore limit our final output $\mathbf{\hat{y}}$ to retrieving from the space of masks generated by SAM: $\mathbf{Y}^K = \{\mathbf{\hat{y}}_k\}_{k=1}^K$. To select the mask that is most representative of the set of generated masks, we consider the compatibility of each mask in the set to every other mask. Concretely, we instantiate this by selecting the mask that has the highest overall image-level similarity to the other masks:
\begin{equation}
    \hat{\textbf{y}} = \argmax_{k=1:K} \mathbb{E}_{\mathbf{\Tilde{y}} \sim p(\textbf{y} \vert \textbf{x})} [ \simm (\mathbf{\hat{y}}_k,\mathbf{\Tilde{y}}) ]
\end{equation}
where for simplicity we use intersection-over-union ($\iou$) to represent our similarity function $\simm(\cdot, \cdot)$ \cite{Monteiro2020StochasticSN}:
\begin{equation}
     \mathbb{E}_{\mathbf{\Tilde{y}} \sim p(\textbf{y} \vert \textbf{x})} [ \simm (\mathbf{\hat{y}}_k,\mathbf{\Tilde{y}}) ] \approx \frac{1}{K} \sum_{k'}^K \iou (\mathbf{\hat{y}}_k,\mathbf{\Tilde{y}}_{k'})
\end{equation}

\section{Experiment}
\subsection{Experimental Setup}

We compare our method principally to the out-of-the-box SAM model \cite{sam}, which previous studies use as a SoTA zero-shot model \cite{medsam-zs, medsam-mcl}, i.e. given input image $\mathbf{x}$, the model predicts $\hat{\mathbf{y}}$ without conditioning on any clicks. We also provide an indication of SAM's upper bound, compared to zero-shot performance, by fine-tuning on the ground-truth annotations for each dataset. We freeze the encoder and update only the decoder's parameters as per \cite{medsam2} and optimise the model using DICE loss. Note, for fair comparison we do not compare with MedSAM \cite{medsam2} as their approach is trained on vast amounts of medical data, and also cannot condition its predictions on user clicks, making it unsuited to our semi-automatic setting.

We evaluate our approach on three publicly available medical imaging datasets.  (1) \textbf{Breast Ultrasound Scan Images (`BUSI')} \cite{AlDhabyani2019DatasetOB} contains 437 images of breast cancer from ultrasound scans (after excluding blank images). (2) \textbf{CVC-ClinicDB (`CVC')} \cite{Bernal2015WMDOVAMF} contains 612 images from 31 colonoscopy sequences for polyps identification. (3) \textbf{ISIC-2016 (`ISIC')} \cite{Gutman2016SkinLA} contains 1279 lesion segmentations from dermoscopic images for identifying melanoma, a lethal form of skin cancer. As we focus on the semi-automatic setting, we provide a bounding box for each image, using the extremity points of the ground-truth mask to mimic an initial user input.  Although our method and main point of comparison are both zero-shot and therefore require no training, to enable fair comparison to the fine-tuned version of SAM, we split the dataset 80 / 10 / 10 into training, validation and test sets respectively, except for ISIC where we use the already split test set and extract 10\% of the train set for validation.

We run our models using the 94M parameter Segment Anything Model (SAM) \cite{sam}, although our method is extensible to future pre-trained segmentation models with similar click-based interaction capabilities. We chose $K = 50$ for top K approximation, which provided a good trade-off between latency and performance. 

Following previous work in medical imaging \cite{medsam}, we evaluate using both Dice Similarity Coefficient (DSC) and Normalised Surface Distance (NSD) to calculate the accuracy of predictions. DSC is a region-based metric that computes the pixel-level harmonic mean between the ground-truth and predicted mask, while NSD is a contour-based metric that computes the consensus between the boundaries of two masks. 

\begin{table*}
\centering
\footnotesize
    \begin{tabular}{llllllll}
\toprule
& & \multicolumn{2}{c}{\textbf{Breast Ultrasound Scan}} & \multicolumn{2}{c}{\textbf{CVC-ClinicDB}} & \multicolumn{2}{c}{\textbf{ISIC-2016}} \\
\cmidrule(lr){3-4} \cmidrule(lr){5-6} \cmidrule(lr){7-8}
\textbf{Section} & \textbf{Method} & \multicolumn{1}{c}{\textbf{DSC} $\uparrow$}  & \multicolumn{1}{c}{\textbf{NSD} $\uparrow$} & \multicolumn{1}{c}{\textbf{DSC} $\uparrow$} & \multicolumn{1}{c}{\textbf{NSD} $\uparrow$} & \multicolumn{1}{c}{\textbf{DSC} $\uparrow$} & \multicolumn{1}{c}{\textbf{NSD} $\uparrow$} \\
\midrule

(A) Fine-tuning & SAM-FT & $88.4 \pm 4.3$ & $52.6 \pm 21.8$ & $94.0 \pm 5.7$ & $81.6 \pm 15.3$ & $94.5 \pm 2.8$ & $56.8 \pm 22.2$ \\
\midrule
\multirow{2}{*}{(B) Zero-shot} & SAM & $79.3 \pm 14.8$ & $38.0 \pm 21.7$ & $86.8 \pm 19.0$ & $64.4 \pm 23.3$ & $\mathbf{81.9 \pm 13.1}$ & $17.4 \pm 17.1$ \\
& \textbf{\textsc{SimSAM} (ours)} & $\mathbf{81.3 \pm 15.2} \ddag$ & $\mathbf{41.6 \pm 21.8} \ddag$ & $\mathbf{87.3 \pm 20.0}$ & $\mathbf{69.2 \pm 23.4} \ddag$ & $81.8 \pm 13.5$ & $\mathbf{19.0 \pm 18.8} \ddag$ \\
\midrule
\multirow{3}{*}{(C) Ablations} & Random $q(\textbf{z} \vert \textbf{x})$ & $83.6 \pm 8.7$ & $42.6 \pm 23.8$ & $82.3 \pm 22.8$ & $62.0 \pm 29.8$ & $69.4 \pm 27.1$ & $21.0 \pm 18.5$ \\
& Pixel aggregation & $77.5 \pm 18.6$ & $39.1 \pm 22.5$ & $82.2 \pm 26.9$ & $65.8 \pm 28.3$ & $78.3 \pm 18.5$ & $17.2 \pm 18.3$ \\
& $K = 1$ & $75.8 \pm 22.2$ & $40.0 \pm 23.4$ & $80.4 \pm 27.8$ & $60.4 \pm 28.8$ & $73.4 \pm 24.6$ & $16.9 \pm 18.6$ \\
\bottomrule
\end{tabular}
    \caption{Mean $\pm$ standard deviation of results on the Breast Ultrasound Scan, CVC-ClinicDB and ISIC-2016 test sets, showing: (A) upper bound obtained from fine-tuning; (B) zero-shot results for our method and the baseline; (C) ablations of key components from our method. \textbf{Bold} indicates best zero-shot result. $\ddag$ indicates statistically significant difference between SimSAM and SAM using Wilcoxon Signed Rank Test ($p < 0.01$).}
    \label{tab:main_results}
\end{table*}

\begin{figure}[t]
\centering
\includegraphics[scale=0.4]{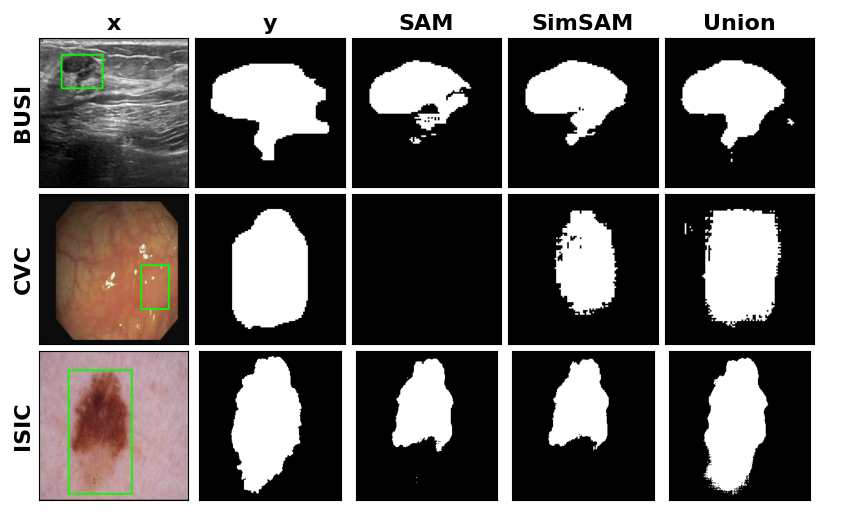}
     \caption{Qualitative results from the Breast Ultrasound Scan, CVC-ClinicDB and ISIC-2016 test sets. (Col. 1) Image $\mathbf{x}$ with bounding box prompt; (Col. 2) binary ground truth mask $\mathbf{y}$; (Col. 3) baseline SAM model \cite{sam}; (Col. 4) \textsc{SimSAM}; (Col. 5) Union of all $K = 50$ masks generated by \textsc{SimSAM}.}
     \label{fig:case_study}
\end{figure}

\vspace{-0.5em}
\subsection{Main Results}
\label{sec:main_results}
Table \ref{tab:main_results}B compares the zero-shot performance of \textsc{SimSAM} to the baseline SAM method. We find consistent outperformance across all three datasets. In terms of DSC, \textsc{SimSAM} obtained comparable or superior performance, showing that even though our approach is originally motivated to improve accuracy around the contours of the image, it does not sacrifice accuracy in regional measures. In terms of NSD, \textsc{SimSAM} obtained consistently significant improvement measured by Wilcoxon Signed Rank Test ($p < 0.01$). This demonstrates our approach was consistently effective at improving accuracy around the region of interest's contours. We found the largest performance gains in the BUSI dataset, which is the more challenging dataset that contains significant uncertainty over edge boundaries. Table \ref{tab:main_results}A provides an indicative upper bound by demonstrating the performance of a fine-tuned version of SAM on each dataset. In terms of latency per sample, \textsc{SimSAM} is only moderately slower at 397ms compared to 245ms for the baseline.

\subsection{Ablation Study}
Table \ref{tab:main_results}C further shows the effect of ablating the key components of our system: (i) in order to verify that the clicks obtained from $q(\textbf{z} \vert \textbf{x})$ provide meaningful information, rather than just adding randomness, we replace the top K clicks from Section \ref{sec:topk} with randomly sampled clicks; (ii) we replace the image-level aggregation module from Section \ref{sec:image-level} with pixel-level averaging across the images; (iii) to investigate the broader benefit from aggregating over multiple masks, we consider the setting where $K = 1$, i.e. we just prompt the model with the most likely click. 

For (i), we find performance declines substantially for CVC and ISIC. As this trend does not hold for BUSI however, we postulate that for BUSI, because the dataset is more challenging, the model is less able to approximate the human distribution over clicks. By contrast, the clicks provided by the random approach are independent and identically distributed across all possible pixels. We find this corroborates findings in other zero-shot tasks such s active learning, which finds random selection to be an effective baseline, when prior knowledge is weak \cite{Liu2023COLosSALAB}. For (ii), we find performance declines across all three datasets, with DSC generally being worse than even the baseline. This supports the importance of accounting for inter-pixel dependencies in the aggregation process. For (iii), we also note a considerable decline, with worse performance across the board compared to the baseline. This shows that simply relying on a single click is more likely to mislead the model, and that it is therefore important to aggregate over multiple possible clicks.  
\vspace{-0.5em}
\subsection{Qualitative Analysis}
\label{sec:case_study}
In Figure \ref{fig:case_study}, we present several qualitative examples to illustrate how \textsc{SimSAM}'s annotations differ visually from SAM. In the top row, we show how \textsc{SimSAM} is often able to repair gaps in the initial annotation, producing an image that is overall smoother and captures more of the ground-truth annotation regions. In the middle row, we show how due to low contrast images, SAM sometimes fails to identify an object entirely, due to its pixel-level probabilities falling below the classification threshold (0.5). By contrast, \textsc{SimSAM}'s aggregation procedure prevents this failure state. Finally, in the bottom row example we see how the ground-truth annotation actually includes a larger region of the tumour beyond what the more obvious edge boundaries would indicate. Although we find both models fail to explicitly capture this, the union region in the right hand column shows that some of \textsc{SimSAM}'s samples were able to capture this. This indicates that there is the potential for further `dark knowledge' about the annotation task to be extracted from the model.
\vspace{-0.5em}

\section{Conclusion and Future Work}
We present \textsc{SimSAM}, a novel extension to SAM for zero-shot medical imaging. We show our method attains SoTA performance across three publicly available datasets, including up to 15.5\% improvement in contour segmentation accuracy. Qualitatively, we demonstrate how our method is able to produce more robust masks that mitigate many of the pitfalls of SAM.

Future work may look to expand the interaction paradigm to sequences of clicks or to include additional forms of interaction such as textual prompts; additional work may consider extending the framework of user input simulation beyond SAM; other work could refine the click simulation, potentially through few-shot learning on real human annotators; finally, as indicated in Section \ref{sec:case_study}, there may be additional dark knowledge that could further be extracted to improve model performance.


\section{Acknowledgments}
\label{sec:acknowledgments}
This work is partly supported by the EPSRC DTP Studentship program. The opinions expressed in this paper are the authors’, and are not necessarily shared/endorsed by their employers and/or sponsors.

\section{Compliance with Ethical Standards}
This research study was conducted retrospectively using human subject data made available in open access. Ethical approval was not required as confirmed by the license attached with the open access data.

\bibliographystyle{IEEEbib}
\bibliography{refs}

\end{document}